# Argumentation as a General Framework for Uncertain Reasoning

John Fox, Paul Krause and Morten Elvang-Gøransson[†]

Advanced Computation Laboratory,
Imperial Cancer Research Fund,
London, UK

†Centre for Cognitive Science,
University of Roskilde,
Denmark

## Abstract

Argumentation is the process of constructing arguments about propositions, and the assignment of statements of confidence to those propositions based on the nature and relative strength of their supporting arguments. The process is modelled as a labelled deductive system, in which propositions are doubly labelled with the grounds on which they are based and a representation of the confidence attached to the argument. Argument construction is captured by a generalised *argument consequence relation* based on the $\wedge$, $\rightarrow$-fragment of minimal logic. Arguments can be aggregated by a variety of numeric and symbolic *flattening functions*. This approach appears to shed light on the common logical structure of a variety of quantitative, qualitative and defeasible uncertainty calculi.

## 1  INTRODUCTION

Bayesian probability theory is the most widely accepted mathematical framework for reasoning under uncertainty. However questions about its universal applicability have often been raised. Before the middle of the seventeenth century, when Pascal and the Port Royal school laid the foundations of the modern theory, frequentistic and numerical ideas about probability were not self-evident, and in the modern period doubts about its generality continue to surface (Hacking 1975). Partly as a response to these doubts various alternative numerical schemes for reasoning under uncertainty, such as fuzzy sets, possibility theory and belief functions, have recently achieved prominence (e.g. Kruse 1991; Kruse and Seigel 1991).

The general assumption that uncertainty must be represented quantitatively has also been widely questioned, particularly in AI. Issues include (a) claims that quantitative notions of uncertainty do not capture important intuitions about doubt and belief, (b) worries about the interpretation of the numbers which should constrain and guide their use, and (c) practical problems such as how to deal with situations where quantitative data are not available. As a consequence interest has grown in the use of qualitative and logical methods for uncertain and defeasible reasoning. Various non-quantitative frameworks for reasoning under uncertainty have been proposed including non-monotonic logics (e.g. Bell 1990), qualitative and semiqualitative calculi (e.g. Kyberg 1991; Wellman 1990), and informal frameworks like Cohen's endorsement scheme (Cohen 1985) and Fox's linguistic predicates (Fox 1984).

There is now a common, though controversial, view that different techniques for managing uncertainty are legitimate and have different contributions to make (Clark, 1990; Fox 1986; Krause and Fox 1991; Saffiotti 1987). Many efforts are under way to develop techniques for using calculi in combination, and also to identify formal frameworks which unify some or all of the proposals (Dubois and Prade 1990; Ginsberg 1988; Krause and Fox 1991; Parsons 1991; Saffiotti and Umkehrer 1991).

Our interest is to develop an extended logic language in which calculi embodying different ideas about uncertainty management can be accommodated in a principled way (Fox et al. 1992). In particular we are exploring the possibility that "argumentation", deducing reasons to believe or doubt propositions (and reasons to believe or doubt arguments) provides a way of doing this. A formal system for argumentation is presented and ways in which this may shed light on a common logical structure for a variety of calculi are discussed.

## 2  ARGUMENTATION

In a standard logic, $L$, an argument is a sequence of inferences leading to a conclusion. If the argument is correct, then the conclusion is true. An argument

$$p_1, ..., p_n \vdash_L q$$

is correct in the logic $L$ if $q$ follows from the rules and the axioms of $L$ augmented with $p_1, ..., p_n$. Therefore a correct argument simply yields a proposition $q$, or a sentence which can be paraphrased as "$q$ is true (in the context $p_1, ..., p_n$)".

In the approach we take, this traditional form of logic based argumentation is extended to allow arguments not only to prove propositions but also merely to indicate support for or even doubt in them, by assigning labels to arguments which designate the confidence warranted by the arguments for their conclusions. Confidences may be expressed in a variety of representations without modifying the underlying inference system.

$$
\begin{array}{ll}
\textbf{Introduction Rules} & \textbf{Elimination Rules}
\end{array}
$$

$(\wedge \text{I})\quad \dfrac{(\varphi, a, S)\quad (\psi, b, S')}{(\varphi \wedge \psi, a \cup b, S \bullet S')}$
$\qquad(\wedge \text{E})\quad \dfrac{(\varphi \wedge \psi, a, S)}{(\varphi, a, S)}\quad \dfrac{(\varphi \wedge \psi, a, S)}{(\psi, a, S)}$

$(\rightarrow \text{I})\quad \dfrac{\overline{(\varphi, a, S)}\;\; \vdots \;\; (\psi, a, S)}{(\varphi \rightarrow \psi, a, S)}$
$\qquad(\rightarrow \text{E})\quad \dfrac{(\varphi, a, S)\quad (\varphi \rightarrow \psi, b, S')}{(\psi, a \cup b, S \bullet S')}$
$\qquad\textbf{Axiom}\quad \dfrac{}{(\varphi, a, S)}$

$(\vee \text{I})\quad \dfrac{(\varphi, a, S)}{(\varphi \vee \psi, a, S)}\quad \dfrac{(\psi, a, S)}{(\varphi \vee \psi, a, S)}$
$\qquad(\vee \text{E})\quad \dfrac{(\varphi \vee \psi, a, S)\quad (\sigma, b', S')\quad (\sigma, b'', S'')}{(\sigma, b' \cup b'', S \bullet (S' \bullet S''))}$

with assumptions $\overline{(\varphi, a, S)}$ and $\overline{(\psi, a, S)}$

$(\neg \text{I})\quad \dfrac{\overline{(\varphi, a, S)}\;\; \vdots \;\; (\bot, a, S)}{(\neg \varphi, a, S)}$
$\qquad(\neg \text{E})\quad \dfrac{(\varphi, a, S)\quad (\neg \varphi, b, S')}{(\bot, a \cup b, S \bullet S')}$

Figure 1: Inference Rules of LA.

## 3 ARGUMENTATION CONSEQUENCE RELATION

Our extended form of argumentation is summarised by the following schema:

Database $\vdash_{ACR}$ (Sentence, Grounds, Sign)

where $\vdash_{ACR}$ is an argumentation consequence relation (ACR). To illustrate, consider the following informal medical argument: *there is reason to believe that the patient has cancer, because she is elderly and has recently lost weight and I know that this is a classical presentation of an advanced malignancy.* Here, the database comprises the beliefs the medical reasoner holds about the patient and the medical theory on which the medical reasoner's justifications are based. The sentence about which we are constructing the argument can be any meaningful locution, in this case: *the patient has cancer.* The grounds are some explicit reflection of the structure of the argument. Ideally the grounds comprise a complete recording of the argument structure, but other ways of representing the grounds are possible, such as the set of assumptions it makes together with an identifier for the theory under which the argument is constructed; the actual choice of representation will depend on the intended application. Finally, the sign is some qualification of the conclusion. In general, signs can be drawn from a variety of "dictionaries" of numerical coefficients, symbolic terms, etc. In the simplest case, the atomic qualifier "supported" might be chosen, meaning that the argument warrants a qualitative increase in belief in the conclusion.

## 4 FLATTENING AND AGGREGATION

Given some set of arguments for $p$ whose confidence labels are drawn from a single dictionary $D_1$ we can *flatten* these arguments to yield a confidence label drawn from $D_2$ to represent the overall confidence in $p$ (we do not exclude the case that $D_1 = D_2$). Suppose the formulae of interest are expressed in the language $L$ and the argument labels belong to $A$. Then a flattening function has the general form:

flat: $\mathbf{P}(L \times A \times D_1) \rightarrow \mathbf{P}(L \times D_2)$

In this paper we are mainly interested in flattening arguments for a specific formula, $p$. We call this *aggregation*:

$\text{flat}_p \colon \mathbf{P}(L \times A \times D_1) \rightarrow D_2$

Aggregation of arguments for and against some proposition means to compute an overall value for the overall confidence in the proposition from the set of individual arguments.

## 5 THE ARGUMENTATION THEOREM PROVER

An argumentation theorem prover (ATP) has been defined in (Krause et al. 1983) and implemented in Prolog which returns propositions labelled with a representation of their supporting arguments. The theorem prover is for an ACR which is generic in the choice of dictionary; it can be *instantiated* with different dictionaries of confidence labels, provided that these satisfy a number of constraints.

A variety of aggregation functions can be defined for each dictionary. The kind of aggregation that results is termed *increasing monotonic aggregation*, because the procedure is only capable of aggregating "positive" support; as we shall see, accommodation of arguments which reduce support for a conclusion requires a little extra machinery.

Using this framework, any argument that can be constructed for a certain proposition from a given set of facts can be computed. In constructing the individual arguments, confidence measures on the facts are propagated to the conclusion of the arguments. Then the overall confidence in the conclusion of the arguments is computed by aggregating the confidence in the individual arguments.

The ATP reported in (Krause et al. 1993) is based on the $\wedge$, $\rightarrow$-fragment of minimal logic with implicit negation (i.e. $\neg p$ is represented by $p \rightarrow \perp$). The results of Ambler (1992) ensures that the construction and aggregation of arguments is done in a mathematical coherent way and (Ambler 1992) provides a formal semantics for increasing monotonic aggregation over arguments constructed using minimal logic. For the labelling of arguments this version of the ATP relies on the strong correspondence between terms of the primitively typed $\lambda$-calculus and proofs in the $\wedge$, $\rightarrow$-fragment of minimal logic. However, here we will label each axiom in a database with (a singleton set containing) an atomic label, and represent an argument by a set of labels referring to the axioms used in constructing that argument.

The confidence labels are drawn from a dictionary with some associated operator $\bullet$, some total order $\leq$ and some top element **1**. Using $p$, $q$ to denote propositions, $a$, $b$ to denote sets of axiom labels and $l$, $m$ to denote confidence labels, we present a general logic of argumentation (LA) in figure 1.

*LA* is generic in the choice of dictionary; for a particular choice, $D$, we use $LA_D$ to denote the resulting argumentation consequence relation.

We will show two examples from (Krause et al. 1993) to show how increasing monotonic aggregation works in practice when applied to arguments in LA.

**Example:** In the first example we choose a discrete dictionary with just two elements:

$$D_{bounded} = \{+,++\},$$

with $++$ denoting absolute certainty and $+$ denoting any weaker form of support. The following operator, minimal support $\bullet$, is defined over $D_{bounded}$:

$$1 \bullet m = m \bullet 1$$
$$+ \bullet 1 = +$$
$$++ \bullet ++ = ++$$

where $l, m \in D_{bounded}$ and $++$ is the top element (i.e. $+ < ++$).

To aggregate the overall confidence in a proposition $p$ given by a database $K$ we need the set of arguments for $p$

that can be constructed from $K$. This set, $(LA_{bnd})^P{}_K$, is defined as:

$$(LA_{bnd})^P{}_K = \{(p, a, I) \mid K \vdash_{LAbnd}(p, a, I)\}$$

The aggregation function, $AGG_{bnd}$ is defined for any database and proposition:

$$AGG_{bnd}(K, p) = \begin{cases} ++, \text{ if } (\exists a)((p, a,++) \in (LA_{bnd})^P{}_K) \\ |(LA_{bnd})^P{}_K|, \text{ otherwise} \end{cases}$$

$AGG_{bnd}$ has the dictionary:

$$D_{boundedNat} = \text{Nat} \cup \{++\},$$

as its co-domain. Let $K$ be a database which captures the knowledge that biological cells are normally growth limited, but tumour cells are not so limited:

$$(cell(x) \rightarrow growthLtd(x), \{c1\}, +)$$
$$(tumourCell(x) \rightarrow cell(x), \{t1\}, ++)$$
$$(tumourCell(x) \rightarrow \neg growthLtd(x), \{t2\}, ++)$$
$$(tumourCell(someX), \{f1\}, ++)$$

Note that formulae with free variables should be read as schemas representing all ground instances of that formula. For simplicity, we have not parameterised the labels with the variables in the corresponding axioms, although this should strictly be done to identify distinct instances of the schemas.

Querying the database $K$ with the question *growthLtd(someX)*, will result in the aggregation function returning the value 1, because there is one supporting argument (whose grounds are $\{c1, t1\}$) and no confirming arguments. Querying with $\neg growthLtd(someX)$, will result in the aggregation function returning the value $++$, because there is one confirming argument (whose grounds are $\{t2, f1\}$).

**Example:** In the second example we assign numerical confidence coefficients to facts in the database and propagate these during the construction of the argument, in a way such that the label of the argument represents a lower bound for the confidence in the conclusion. The dictionary is:

$$D_{num} = [0,1]$$

with 1 as the top element and multiplication as the distinguished operator.

The aggregation function is a generalization of Bernoulli's rule, (Bernoulli 1713). Let $K'$ be the database backing the informal medical argument given above regarding the elderly patient, $X$, with weight loss. Consider the two arguments, informally written as:

$$(cancer, \text{"X has weightloss"}, 0.7)$$
$$(cancer, \text{"X is elderly"}, 0.5)$$

For the sake of the example the symptoms are treated as independent and the numerical coefficients have been chosen arbitrarily. Aggregating in accordance with Bernoulli's rule yields:

$$AGG_{num}(K', cancer) = 0.7+0.5-(0.7 \times 0.5) = 0.85,$$

where $AGG_{num}$ represents Bernoulli's rule. (A more elaborate version of this example in which dependencies between arguments are considered is given in (Krause et al. 1993).)

Essentially arguments in LA have the role of computing "atomic" pieces of support that can be constructed in the context of some database. From the set of all these atomic pieces the overall confidence can be aggregated. In calculating the confidence in a specific proposition, evidence for its negation is not taken into account. However, the concept of flattening allows for such heuristics to be added on top of the basic notion of aggregation.

Because the ATP works by computing atomic pieces of support, a formal correspondence to certain aspects of the quantitative uncertainty calculi can be established. An informal example of this has already been given in the above example using Bernoulli's rule. In (Ambler 1992), a formal correspondence to the Dempster-Shafer theory has been established.

## 6   AGGREGATION CRITERIA

Based on abstract dictionaries we define a set of constraints, that can be used to classify the different kinds of aggregation. The idea is that a specific kind of aggregation satisfies a set of constraints if its dictionary can be projected onto the abstract dictionary in a way such that the constraint is satisfied.

The simplest abstract dictionary is:

$$D_{generic} = \{+\}.$$

An ACR based on this dictionary takes any argument for a proposition as a supporting argument, that is it results in an overall increase in the confidence in the proposition. For such calculi, therefore, there is a constraint on the way in which the relative force of sets of arguments are measured:

$$\mathrm{flat}_p(Arg) \leq \mathrm{flat}_p(Arg \cup \{(p, a, +)\}) \quad (F1)$$

Here, $Arg$ is any set of (Proposition, Grounds, Sign) argument triples for the proposition p. The weak constraint (F1) is satisfied by any instantiation of the ATP. In this 'generic' calculus arguments have equal force; no 'belief mass' is assigned to them.

In general there is no limit to the number of arguments that can be constructed for a proposition. Intuitively, however, some arguments are conclusive. This is captured by the abstract dictionary:

$$D_{bounded} = \{+,++\}.$$

In the presence of a conclusive argument flattening is completely dominated by this argument:

$$\mathrm{flat}_p(\{(p, a, ++)\}) = \mathrm{flat}_p(Arg \cup \{(p, a, ++)\}) \quad (F2)$$

This constraint is also satisfied by any instantiation of the ATP, by mapping the top element 1 of the dictionary to ++.

So far we have only been concerned with increase in confidence. We may define a "complementation operator", -, as: if $p = \neg q$, then $-p = q$, otherwise $-p = \neg p$. Then, the complementary possibility of having a decrease in confidence is reflected by the dictionary

$$D_{delta} = \{+,-\}.$$

In order to strengthen the generic calculus by representing the relationship between $p$ and $-p$ the complementation constraint (C1) is defined to explicate the relationship between a proposition and its dual.

$$\vdash_{ACR} (p, a, +) \mathrm{\ iff} \vdash_{ACR} (-p, a, -) \quad (C1)$$

This constraint is on the ACR and not on the flattening function. It is generic in the definition of duality, although this will normally coincide with logical negation. The only modification that can be allowed is a strengthening of logical negation, reflected in the requirement (C2).

$$\vdash_{ACR} (\neg p, a, +) \mathrm{\ if} \vdash_{ACR} (-p, a, +) \quad (C2)$$

Flattening in a delta calculus should honour the constraint (F1) as well as (F3) below.

$$\mathrm{flat}_p(Arg) \geq \mathrm{flat}_p(Arg \cup \{(p, a, -)\}) \quad (F3)$$

Interestingly, the instantiations of the ATP do not satisfy (C1) and they only satisfy (F3) because $\geq$ coincides with =. In aggregating arguments for (against) a proposition, any argument against (for) this proposition is simply neglected. However, this defect can be circumvented by explicitly aggregating both arguments pro and con and computing the overall confidence on the basis of these two individual confidences.

Finally we specialise the delta calculus with an extended dictionary,

$$D_{bounded\text{-}delta} = \{++, +, -, --\}$$

including symbols for upper and lower confidence bounds. A bounded delta-calculus must satisfy the above constraints together with:

$$\vdash_{ACR} (p, a, ++) \mathrm{\ iff} \vdash_{ACR} (-p, a, --) \quad (C3)$$

$$\vdash_{ACR} (\neg p, a, ++) \mathrm{\ if} \vdash_{ACR} (-p, a, ++) \quad (C4)$$

$$\mathrm{flat}_p(\{(p, a, --)\}) = \mathrm{flat}_p(Arg \cup \{(p, a, -)\}) \quad (F4)$$

But note that constraint F4 does not apply if Arg contains a confirming argument for p (i.e $\exists b.(p, b, ++) \in Arg$); this would correspond to a flat contradiction. Still more specific abstract dictionaries can be introduced to give finer characterizations of specific kinds of aggregation, but the above four calculi provide an appropriate starting point.

## 7   QUANTITATIVE UNCERTAINTY CALCULI

We suggest that argumentation can provide a common logical form for different uncertainty calculi. As the examples have revealed, aggregation of arguments is a hybrid of qualitative and quantitative reasoning. The aggregation criteria were mainly designed to reflect properties of qualitative reasoning, but in a naive way they also reflect prop-erties of quantitative calculi with conditional confidence values.

For instance, if we assume $p$ and the theory we are using includes the causal relationship, $P(q|p, \Delta) > P(q| \Delta)$, then as in simple probability theory our confidence in $q$ will increase in the context $\Delta$. A way of making this increase in confidence explicit over a dictionary consisting of the interval $[0, \infty]$ is by computing the ratio:

$$\frac{P(q| \Delta)}{P(- q| \Delta)}$$

In the particular example above, we would then see the increase as:

$$\frac{P(q| p, \Delta)}{P(- q| p, \Delta)} \qquad p \notin \Delta$$

# 8   ARGUMENTATION AND DEFEASIBILITY

In general there are no consistency requirements on the databases we accept. For instance in the bounded-delta calculus we may derive $(p, a, ++)$ and $(p, a', --)$, where $a$, $a'$ are different labels. (If the argument labelling algebra is not strict enough, then it can even happen that $a = a'$, but such ACRs would be too weak to reflect the essential feature of most uncertainty calculi: the counting up of independent pieces of evidence.) This ability to represent contradictory conclusions may at first appear to be an unacceptable state of affairs, but it is in fact highly desirable. Situations where one can claim to be able to argue for some proposition $p$ while also being able to argue against $p$ on different assumptions are commonplace. In medicine, for instance, opinions which are flatly contradictory or inconsistent with observations or received wisdom are not unusual. Scientific progress actually depends upon competing theories, with new theories often arising from anomalies, contradictory findings, and observed exceptions to well entrenched beliefs. Since practical reasoning cannot avoid inconsistencies the ability to manage them is important (e.g. Fox et al. 1991a; Gabbay and Hunter 1991).

Conflicts caused by such inconsistencies can be resolved by considering the logical certainty of the arguments. There are several approaches that can be taken, but we will restrict the present discussion to *discounting* and *rebutting* of arguments in the bounded-delta calculus.

## 8.1   DEFEATING AN ARGUMENT

More advanced uses of arguments emerge when we allow arguments to have different degrees of logical certainty, made explicit through a relation on the argument labels. One step in this direction is to allow one argument to defeat another.

An argument $(P, a, l)$, $l \in \{+,++\}$, is discounted by the

argument $(Q, b, m)$, $\in \{-,--\}$, if the argument labelled a depends on $Q$.

An argument $(P, a, l)$, $\in \{+,++\}$, is rebutted by the argument $(Q, b, m)$, $\in \{-,--\}$, if $P = Q$ and either $m = --$ or $m = -$ and $l = +$.

An argument $(P, a, l)$ is defeated, if it is discounted or rebutted by some argument which is not defeated.

The concepts of rebutting and discounting are very similar to Pollock's (1992) notions of rebutting and undercutting defeat, though our work is based on that of Toulmin (1958). Toulmin's work was not formal in a mathematical sense. However, he did provide a useful schematic representation of an argument. Essentially, he modelled an argument as a claim supported by some data, with that claim backed by a warrant. A claim may be rebutted by an argument that directly refutes the claim, though we extend this with the notion of discounting an argument by attacking the validity of the data or warrant in the given context.

## 8.2   SELECTIVE AGGREGATION

A new form for aggregation can now be defined, where only undefeated arguments are taken into account. This form of aggregation depends on a hybrid of object-level and logical certainty assigned to the individual arguments. Object-level certainty is assigned as confidence labels and logical certainty is determined through the notion of defeat. Put simply, undefeated arguments are more certain than defeated ones. A full discussion of this approach to logical certainty can be found in (Elvang-Gøransson et al. 1993).

# 9   META-ARGUMENTATION

In previous presentations this more advanced form for aggregation was called Meta-argumentation, essentially reflecting that arguments are themselves objects of arguments. The ambition of the Meta-argumentation program was to make argumentation structures explicit as objects in a meta-logic of argumentation and supply expressive power for representing discounting and rebutting in the language itself. This allowed for the further extension of making conflicts explicit in this meta-language. In this presentation we have decided to accommodate these concepts in the flattening functions, resulting in a more coherent presentation within the realm of LDS.

# 10   RESOLVING CONFLICTS AND MAKING DECISIONS

So far, arguments have been backed only by a single theory. Therefore conflicts were essentially caused by logical inconsistencies in the theory. Only through the extra structure added to databases and arguments could such conflicts occasionally be resolved. Also, the aggregation procedures have been based on aggregation over any constructible argument. However, it is not always the case that arguments are constructed from the same database and quite often decisions are made from an incomplete set of argu-ments that are regarded as sufficient to back the decision.

One interesting approach would be to view arguments as constructed contingently on an agent's database. Thus, for example, if two agents with conflicting databases are debating a proposition, then this might lead to conflicting arguments being formulated. This kind of conflict is not caused by logical inconsistency in a particular theory, but rather by disagreement over what data are applicable to the problem at hand; the conflict can only be resolved by making at least one of the agents revise its theory. Analogously, a decision maker may be thought of as engaging in a private debate, in which conflicting arguments are formulated on different hypotheses for interpreting the data. Such agents can be conceived as having reflective capabilities, acknowledging that their beliefs (theories) can lead to contradictory conclusions and if such contradictions are detected one or more theories will need to be modified.

There are a number of strategies which may be taken to resolve such conflicts, the precise strategy taken depending on the nature of the situation at hand. A more extensive account of conflict resolution strategies within the framework of argumentation can be found in (Elvang-Gøransson et al. 1993).

## 11  CONCLUSION

This has been a wide ranging survey of the use of argumentation as a general purpose framework for reasoning under uncertainty. For brevity, much of the discussion has been kept relatively informal, but most of the work discussed has received rigorous development; the main results are to be found in (Ambler 1992), (Elvang-Gøransson 1993) and (Krause et al. 1993).

The basic ideas of argumentation have been in practical use in medical decision making applications for some time (Fox et al. 1991b; Huang et al. 1993). The formal developments presented here validate these methods, and suggest important extensions to the practical techniques currently in use. Although our applications have so far been confined to medical decision making, the formal results suggest that an argumentation framework offers a general basis for developing reasoning techniques which accommodate quantitative, qualitative and wider logical approaches to the management of uncertainty.

### Acknowledgements

Paul Krause is supported under the DTI/SERC project number 1822: a Formal Basis for Decision Support Systems. The authors wish to thank Simon Ambler and Michael Clarke of Queen Mary and Westfield College, London, for their many contributions in the development of these ideas. We would also like to thank the anonymous referees for valued comments.